\documentclass[10pt,twocolumn,letterpaper]{article}

\usepackage{iccv}
\usepackage{times}
\usepackage{epsfig}
\usepackage{graphicx}
\usepackage{amsmath}
\usepackage{amssymb}

\usepackage{comment}
\usepackage[inline]{enumitem}
\usepackage{svg}
\usepackage{multirow}
\usepackage{epstopdf}
\usepackage[export]{adjustbox}

\renewcommand{\vec}[1]{\mathbf{#1}}
\usepackage{makecell}
\renewcommand\arraystretch{1.2}

\usepackage[breaklinks=true,bookmarks=false]{hyperref}

\iccvfinalcopy 

\def\iccvPaperID{2256} 

\ificcvfinal\fi
\begin{document}

\title{Non-linear Convolution Filters for CNN-based Learning}

\author{Georgios Zoumpourlis \quad Alexandros Doumanoglou \quad Nicholas Vretos \quad Petros Daras\\
Information Technologies Institute, Center for Research and Technology Hellas, Greece\\
6\textsuperscript{th} km Charilaou-Thermi Road, Thessaloniki, Greece\\
{\tt\small \{{zoump.giorgos, aldoum, vretos, daras}\}@iti.gr}
}

\maketitle
\thispagestyle{empty}

\begin{abstract}
   During the last years, Convolutional Neural Networks (CNNs) have achieved state-of-the-art performance in image classification. Their architectures have largely drawn inspiration by models of the primate visual system. However, while recent research results of neuroscience prove the existence of non-linear operations in the response of complex visual cells, little effort has been devoted to extend the convolution technique to non-linear forms. Typical convolutional layers are linear systems, hence their expressiveness is limited. To overcome this, various non-linearities have been used as activation functions inside CNNs, while also many pooling strategies have been applied. We address the issue of developing a convolution method in the context of a computational model of the visual cortex, exploring quadratic forms through the Volterra kernels. Such forms, constituting a more rich function space, are used as approximations of the response profile of visual cells. Our proposed second-order convolution is tested on CIFAR-10 and CIFAR-100. We show that a network which combines linear and non-linear filters in its convolutional layers, can outperform networks that use standard linear filters with the same architecture, yielding results competitive with the state-of-the-art on these datasets.
\end{abstract}

\section{Introduction}

     Convolutional neural networks (CNNs) have been shown to achieve state-of-the-art results on various computer vision tasks, such as image classification. Their architectures have largely drawn inspiration by models of the primate visual system, as the one described by Hubel and Wiesel \cite{Hubel62}. The notion of convolution, used to mimic a functional aspect of neurons in the visual cortex, is critical to understand their success.
     
     Typical convolutional layers are linear systems, as their outputs are affine transformations of their inputs. Due to their linear nature, they lack the ability of expressing possible non-linearities that may actually appear in the response of complex cells in the primary visual cortex \cite{Rapela06}. Hence, we claim that their expressiveness is limited. To overcome this, various non-linearities have been used as activation functions inside CNNs, while also many pooling strategies have been applied. Little effort has been devoted to explore new computational models that extend the convolution technique to non-linear forms, taking advantage of the research results of neuroscience, that prove the existence of non-linear operations in the response of visual cells \cite{Szulborski90}\cite{Niell08}. The complexity of human visual cortex demonstrates gaps that need to be bridged by CNNs, regarding the way convolution operations are applied. One of these gaps, is the exploration of higher-order models.
     
     In this work, we study the possibility of adopting an alternative convolution scheme to increase the learning capacity of CNNs by applying Volterra's theory \cite{Volterra30}, which has been used to study non-linear physiological systems, adapting it to the spatial domain. Considering the convolution operation, instead of summing only linear terms to compute a filter's response on a data patch, we propose to also sum the non-linear terms produced by multiplicative interactions between all the pairs of elements of the input data patch. Transforming the inputs through a second-order form, we aim at making them more separable. In this way, convolution filters with more rich properties in terms of selectivity and invariance are created.
     
     The novelties of the proposed work are:
     \begin{itemize}
        \item The incorporation of a biologically plausible non-linear convolution scheme in the functionality of CNNs
        \item The derivation of the equations that describe the forward and backward pass during the training process of this filter type
        \item A CUDA-based implementation of our method as a non-linear convolutional layer's module in Torch7\footnote{http://torch.ch/}\cite{torch7}
     \end{itemize}

     The rest of the paper is organized as follows: in Section \ref{RelWork}, related work is outlined. In Section \ref{PropMeth}, the proposed method is described, theoretically grounded to Volterra's computational method, and the concept of training is mathematically explained, while a description of our scheme's practical implementation is given in Section \ref{CodeImpl}. In Section \ref{Experiments} experimental results on CIFAR-10 and CIFAR-100 datasets are drawn and finally in Section \ref{Conclusion} the paper is concluded.


\section{Related Work}
\label{RelWork}

     One of the first biologically-inspired neural networks, was Fukushima's Neocognitron \cite{Fukushima80}, which was the predecessor of CNN, as it was introduced by LeCun \textit{et~al.} in \cite{LeCun90}. Convolutional layer is the core building block of a CNN. Early implementations of CNNs have used predefined Gabor filters in their convolutional layers. This category of filters can model quite accurately the properties of simple cells found in the primary visual cortex (V1) \cite{Marcelja80}.
     
     This type of visual cell has a response profile which is characterized by spatial summation within the receptive field. Finding the optimal spatial stimuli \cite{Foldiak01} for simple cells is a process based on the spatial arrangement of their excitatory and inhibitory regions \cite{Movshon78}. However, this does not hold true for complex visual cells. Also, we cannot obtain an accurate description of their properties, by finding their optimal stimulus.
     
     This fact has been ignored by most of the CNN implementations so far, as they have settled to the linear type of convolution filters, trying to apply quantitative rather than qualitative changes in their functionalities.  He \textit{et~al.} \cite{He15} proposed Residual Networks (ResNets), which have shortcut connections parallel to their normal convolutional layers, as a solution to the problems of vanishing/exploding gradient and hard optimization when increasing the model's parameters (i.e. adding more layers).
     Zagoruyko \& Komodakis \cite{WRN16} showed that wide ResNets can outperform ResNets with hundrends of layers, shifting the interest to increasing the number of each layer's filters. Alternatively to works that focus on creating networks with more convolutional layers or more filters, we evaluate the impact of using both non-linear and linear terms as approximations of convolution kernels to boost the performance of CNNs.
     
     Apart from ResNets, very low error rates have also been achieved in the ImageNet Challenge \cite{Russakovsky2015} by methods that used their convolutional layers in new ways, enhancing their representation ability. Lin \textit{et~al.} \cite{Nin14} proposed \text{``}Network in Network (NIN)\text{''}, as a remedy to the low level of abstraction that typical filters present. Instead of the conventional convolution filter, which is a generalized linear model, they build micro neural networks with more complex structures to abstract the data within the receptive field. To map the input data to the output, they use multilayer perceptrons as a non-linear function approximator, which they call \text{``}mlpconv\text{''} layer. The output feature maps are obtained by sliding the micro networks over the input in a similar manner as CNN. Szegedy \textit{et~al.} \cite{Szegedy15} introduced a new level of organization in the form of the \text{``}Inception module\text{''}, which uses filters of variable sizes to capture different visual patterns of different sizes, and approximates the optimal sparse structure. Xie \textit{et~al.} \cite{ResNeXt16} proposed a way to exploit the split-transform-merge strategy of \text{``}Inception\text{''} models, performing a set of transformations, each on a low-dimensional embedding, whose outputs are aggregated by summation.
     
     The authors of \cite{RCNN15}, based on the abundancy of recurrent synapses in the brain, proposed the use of a recurrent neural network for image classification. They proved that inserting recurrent connections within convolutional layers, gives boosted results, compared to a feed-forward architecture. Their work is a biologically plausible incorporation of mechanisms originating from neuroscience into CNNs.
     
	In \cite{Sejnowski86}, a Boltzmann learning algorithm is proposed, where feature interactions are used to turn hidden units into higher-order feature detectors. In \cite{SHBM14}, an efficient method to apply such learning algorithms on higher-order Boltzmann Machines was proposed, making them computationally tractable for real problems.
     
     In \cite{Bengio09}, Bergstra \textit{et~al.} created a model for neural activation which showed improved generalization on datasets, by incorporating second-order interactions and using an alternative non-linearity as activation function.
      
     In \cite{Analysis06}, an attempt is made to analyze and interpret quadratic forms as receptive fields. In their study, it was found that quadratic forms can be used to model non-linear receptive fields due to the fact that they follow some of the properties of complex cells in the primary visual cortex. These properties include response to edges, phase-shift invariance, direction selectivity, non-orthogonal inhibition, end-inhibition and side-inhibition. In constrast to the standard linear forms, in quadratic and other non-linear forms the optimal stimuli do not provide a complete description of their properties. It is shown that no invariances occure for an optimal stimulus while for other general sub-optimal stimuli there may exist many invariances which could be of a large number but lack easy interpretation. Although the optimal stimulus is not related to a filter's invariance, its neighborhood is studied under a more loose sense of transformation invariance. It is shown that proper quadratic forms can demonstrate invariances in phase-shift and orientation change. From the previous discussion we conclude that using non-linear forms to convolutional layers may be a reasonable future direction in computer vision.


\section{Proposed Method}
\label{PropMeth}

     The proposed method, as earlier stated, makes use of the Volterra kernel theory to provide means of exploiting the non-linear operations that take place in a receptive field. Up to now, and to the best of our knowledge, non-linearities were exploited mainly through the activation functions and pooling operations between different layers of CNNs. Nevertheless, such non-linearities may be an approach to code inner processes of the visual system, but not the ones that exist in a receptive field's area.

     Our method follows the typical workflow of a CNN, by lining up layers of different purposes (convolution, pooling, activation function, batch normalization, dropout, fully-connected etc.), while a non-linear convolutional layer can be plugged in practically in all existing architectures. Nevertheless, due to its augmentation of trainable parameters involved, care should be taken for the complexity of the overall process. To that end, a CUDA implementation in Section \ref{CodeImpl} is also provided.

       \subsection{Volterra-based convolution}
       
     The Volterra series model is a sequence of approximations for continuous functions, developed to represent the input-output relationship of non-linear dynamical systems, using a polynomial functional expansion. Their equations can be composed by terms of infinite orders, but practical implementations based on them use truncated versions, retaining the terms up to some order $r$.
     
      In a similar way to linear convolution, Volterra-based convolution uses kernels to filter the input data. The first-order Volterra kernel, contains the coefficients of the filter's linear part. The second-order kernel represents the coefficients of quadratic interactions between two input elements. In general, the $r$-th order's kernel represents the weights that non-linear interactions between $r$ input elements have on the response. In the field of computer vision, Volterra kernels have been previously used in \cite{Volterrafaces} for face recognition, serving effectively as approximations of non-linear functionals.

     \subsection{Forward pass}
       
     For our proposed convolution, we adopted a second-order Volterra series. Given an input patch $\vec{I}\in{\rm I\!R}^{k_h \times k_w}$ with $n$ elements ($n=k_h\cdot k_w$), reshaped as 
a vector $\vec{x}\in{\rm I\!R}^{n}$:
\begin{equation}
\vec{x} = { \begin{bmatrix} x_1 & x_2 & \cdots & x_n \end{bmatrix} }^T
\end{equation}
the input-output function of a linear filter is:
    \begin{equation}	\label{eq_fp_lin}
       y(\vec{x}) = 
	      \sum_{i=1}^{n}{
	          \big(w_1^ix_i\big)} 
	    + b
    \end{equation}
    
    where $w_1^i$ are the weights of the convolution's linear terms, contained in a vector $\vec{w}_1$, and $b$ is the bias. In our approach, this function is expanded in the following quadratic form:
     
	\begin{equation}	\label{eq_fp}
       y(\vec{x}) = 
	      \sum_{i=1}^{n}{
	          \big(w_1^ix_i\big)} 
	    + \sum_{i=1}^{n}{\sum_{j=i}^{n}{
	          \big(w_2^{i,j} x_ix_j \big)}}
	    + b
    \end{equation}       

where $w_2^{i,j}$ are the weights of the filter's second-order terms. To avoid considering twice the interaction terms for each pair of input elements $(x_i,x_j)$, we adopt an upper-triangular form for the matrix $\vec{w}_2$ containing their weights, so that the number of trainable parameters for a second-order kernel is $n(n+1)/2$. The generic type to compute the total number of parameters, $n_V$, for a Volterra-based filter of order $r$ is:
       \begin{equation}
       \label{eq_nv}
       n_V = \frac{(n + r)!}{n! r!}
       \end{equation}
       
In a more compact form, (\ref{eq_fp}) is written as:
       \begin{equation}
       y(\vec{x}) = \underbrace{ \vec{x}^T \vec{w}_2 \vec{x} }_\text{quadratic term} + \underbrace{ {\vec{w}_1}^T \vec{x} }_\text{linear term} + b
       \end{equation}
       
       while for the Volterra kernels we have:

       \begin{equation}
       \vec{w}_2 = \begin{bmatrix}
                     w_2^{1,1} & w_2^{1,2} & \cdots & w_2^{1,n} \\ 
                     0         & w_2^{2,2} & \cdots & w_2^{2,n} \\ 
                     \vdots    & \vdots    & \ddots  & \vdots\\ 
                     0         & 0         & \cdots  & w_2^{n,n}
       \end{bmatrix}
       \end{equation}

containing the coefficients $w_2^{i,j}$ of the quadratic term, and:

       \begin{equation}
       {\vec{w}_1}^T = \begin{bmatrix}
                                w_1^1 & w_1^2 & \cdots & w_1^n
       \end{bmatrix}
       \end{equation}

containing the coefficients $w_1^i$ of the linear term. The proposed convolution's output can thus be rewritten as:

       \begin{equation}
       y(\vec{x}) =
       \begin{bmatrix}
       w_2^{1,1}\\ w_2^{1,2}\\ w_2^{1,3}\\ \vdots \\ w_2^{n,n}
       \end{bmatrix}^T
       \begin{bmatrix}
       x_1x_1 \\ x_1x_2 \\ x_1x_3 \\ \vdots \\ x_nx_n
       \end{bmatrix}
       +
       \begin{bmatrix}
       w_1^1 \\ w_1^2 \\ w_1^3 \\ \vdots \\ w_1^n
       \end{bmatrix}^T
       \begin{bmatrix}
       x_1 \\ x_2 \\ x_3 \\ \vdots \\ x_n
       \end{bmatrix}
       + b
       \end{equation}
       
Note that superscripts $(i,j)$ to weights $w_2^{i,j}$ denote correspondence to the spatial positions of the input elements $x_i$ and $x_j$ that interact.

\subsection{Backward pass}
\label{Sect:bp}
       
The derivation of the equations for the backward pass of the Volterra-based convolution, is done by adapting the classic backpropagation scheme to the aforementioned input-output function of (\ref{eq_fp}). To train the weights of the Volterra kernels, we need to compute the gradients of the layer's output $y(\vec{x})$, with respect to the weights $w_1^i$ and $w_2^{i,j}$. To propagate the error, we have to compute the gradients of the layer's output $y(\vec{x})$, with respect to the inputs $x_i$. Hence, $\frac{\partial y}{\partial w_1^i}$, $\frac{\partial y}{\partial w_2^{i,j}}$ and $\frac{\partial y}{\partial {x_i}}$ are the terms that will be used to optimize the weight parameters of our Volterra-based convolutional layer and minimize the network loss. The mathematical equations of backpropagation, are as follows:

\begin{equation}
\label{eq_bp_w}
  \frac{\partial y}{\partial w_1^i} = x_i \qquad\text{ }\qquad \frac{\partial y}{\partial w_2^{i,j}} = x_ix_j
\end{equation}

\begin{equation}
\label{eq_bp_x}
  \frac{\partial y}{\partial x_i} = w_1^i + \sum_{k=1}^{i}{\big( w_2^{k,i} x_k \big)} + \sum_{k=i}^{n}{\big( w_2^{i,k} x_k \big)}
\end{equation}


\section{Quadratic convolution filter implementation}
\label{CodeImpl}

     In order to experiment with the non-linear convolution filters, we used the Torch7 scientific framework. Volterra-based convolution was implemented as a module integrated with the CUDA backend for the Neural Network (cunn) Package of Torch7. Writing a module in Torch7 mainly consists of implementing the module's forward pass (\ref{eq_fp}) as well as the computation of the module's gradients ($\frac{\partial E}{\partial {\vec{w}}}$ and $\frac{\partial E}{\partial {\vec{x}}}$), that are used in back-propagation. We denote by $E$ the error defined by the network's criterion function and refer to $\frac{\partial E}{\partial {\vec{w}}}$ as the layer's weight gradient and $\frac{\partial E}{\partial {\vec{x}}}$ as the layer's input gradient. To implement the forward pass in CUDA, we used the standard im2col \cite{Chellapilla} pattern to unfold data patches into columns, followed by a matrix multiplication with the Volterra-based filter weights. The im2col operation is conducted in parallel by a CUDA kernel, while for the matrix multiplication we used the well established CUDA BLAS functions. Subsequently, computing the weight gradients is, to some extent, similar to computing the forward pass. Once again, the im2col operation is executed on the input image as a CUDA kernel and its output matrix is multiplied with the previous layer's input gradient resulting into $\frac{\partial E}{\partial {\vec{w}}}$. The most expensive operation in a Volterra-based convolutional layer is the computation of the input gradients. As already mentioned before, in contrast to linear convolution, where the input gradient is independent of the provided input, our layer's  input gradient is input-dependent. Thus, to compute the matrix of input gradients, firstly we compute an unfolded matrix containing the gradients of the output with respect to the input. This matrix is then multiplied with the previous layer's input gradient using CUDA BLAS functions. Finally, an appropriate inverse col2im CUDA kernel aggregate operation results in the final matrix of the Volterra-based layer's input gradients $\frac{\partial E}{\partial \vec{x}}$.

     A major difference between the proposed convolution scheme and linear convolution, is the fact that in our case $\frac{\partial y}{\partial {x_i}}$ is a function dependent on $x_i$. This means that, in contrast to standard filters, this term is different for every single patch of a feature map, resulting in an extra computational cost, when the error must be propagated to preceding trainable layers in the network. This cost is proportionate to $H_o \cdot W_o$, where $H_o$ and $W_o$ are the height and the width of the layer's output feature map, respectively. Our layer's code is available at \url{http://vcl.iti.gr/volterra}.
     
\section{Experiments}
\label{Experiments}

\begin{table*}[ht]

\centering
\setlength{\tabcolsep}{.3em}
\renewcommand\arraystretch{1.3}
\begin{tabular}{c|c|c}

\textbf{Network stage} & \textbf{Output size} & \textbf{Model} (d=28, N=4, k=10) \\

\Xhline{3\arrayrulewidth}
\multirow{2}{*}{Initial convolution} & \multirow{2}{*}{$32\times32$} & Batch Normalization \\
 &  & Conv $3\times3$, 16 $\cdot k$ \\

\hline
\multirow{2}{*}{Group 1} & \multirow{2}{*}{$32\times32$} & \multirow{2}{*}{Conv $\begin{bmatrix}3\times3, 16 \cdot k \\ 3\times3, 16 \cdot k \end{bmatrix}$ $\times N$ blocks} \\
  &  &  \\

\hline
\multirow{2}{*}{Group 2} & \multirow{2}{*}{$16\times16$} & \multirow{2}{*}{Conv $\begin{bmatrix}3\times3, 32 \cdot k \\ 3\times3, 32 \cdot k \end{bmatrix}$ $\times N$ blocks} \\
  &  &  \\

\hline
\multirow{2}{*}{Group 3} & \multirow{2}{*}{$8\times8$} & \multirow{2}{*}{Conv $\begin{bmatrix}3\times3, 64 \cdot k \\ 3\times3, 64 \cdot k \end{bmatrix}$ $\times N$ blocks} \\
  &  &  \\

\hline

\multirow{3}{*}{Pooling} & $8\times8$ & Batch Normalization \\
 & $8\times8$ & ReLU  \\
 & $1\times1$ & Average Pooling, $8\times8$ \\
\hline
\multirow{2}{*}{Classifier} & & Fully-Connected $10 (100)$  \\
 & & SoftMax $10 (100)$ \\

\hline
\end{tabular}
\caption{ Network architecture used in our experiments.}
\label{table:Volterra-architectures}
\end{table*}


     We measure the performance of our proposed Volterra-based convolution on two benchmark datasets:
     
     CIFAR-10 and CIFAR-100 \cite{Kriz09}, running our experiments on a PC equipped with  Intel i7-5820K CPU, 64GB RAM and Nvidia Titan X GPU. The Volterra-based convolutional layer was implemented in Torch7. We first describe the experimental setup, then we show a quantitative analysis, in terms of parameters, classification error and train loss, for the proposed method.

\subsection{CNN architecture selection}

   As explained in Section \ref{CodeImpl}, using the proposed convolution in multiple layers of a CNN, an extra computational overhead is introduced during backpropagation. For this reason, we restrain ourselves to testing such filters only in the first convolutional layer of a CNN model. We choose the modern architecture of Wide ResNet \cite{WRN16}, which mainly consists of a convolutional layer, followed by 3 convolutional groups and a classifier. If $d$ is such a network's depth, then each convolutional group contains
   
   $N=(d-4)/6$
   convolutional blocks. In a group, the number of each convolutional layer's filters, is controlled by the widening factor $k$. In our architecture, we follow the above rules, making three changes:
\begin{enumerate*}[label={\alph*)}]
\item we insert a Batch Normalization layer in the start of the network
\item we change the number of the first convolutional layer's output channels, from $16$ to $16 \cdot k$ (i.e., equal to the number of the first group's output channels) and
\item we change the shortcut of the first block in the first group, into an identity mapping, as a consequence of our second change.
\end{enumerate*}
The first change is crucial to prevent the output of the Volterra-based convolution from exploding, due to the multiplicative interaction terms $x_ix_j$. In our experiments, parameter $\gamma$ of the affine transformation $y=\gamma\hat{x}+\beta$ that is applied in this layer, settles to values $0<\gamma<1$. The second change was chosen so that, when the Volterra-based convolution is applied in the first convolutional layer, there are enough non-linear filters to be learnt, producing a feature-rich signal. The third change is done because when a block's input and output channels are equal, then its shortcut is an identity mapping, so that its input is added to its output, without the need to adjust the feature channels in the shortcut by using a convolutional layer. In this case, the signal of the first convolutional layer flows intact through the shortcuts of the first group's blocks.

   The model used in our experiments is described in Table \ref{table:Volterra-architectures}.
   To evaluate the impact of applying the Volterra-based convolution on each dataset, we tested two versions of the general CNN model. The first version, which serves as the baseline model, does not use any non-linear convolution filter. The other version contains non-linear filters in the first convolutional layer and linear filters in all the convolutional groups of the network. 

\begin{table}[b]
\centering
\setlength{\tabcolsep}{.3em}
\begin{tabular}{|c|c|c|}
\hline
\textbf{Epoch} & \textbf{Learning rate} & \textbf{Weight decay} \\
\hline
$1-59$    & $0.1$    & $0.0005$ \\ \hline
$60-119$  & $0.02$   & $0.0005$ \\ \hline
$120-159$ & $0.004$  & $0.0005$ \\ \hline
$160-199$ & $0.0008$ & $0.0005$ \\ \hline
$200-220$ & $0.0008$ & $0$    \\
\hline
\end{tabular}
\caption{Learning rate and weight decay strategy used in our experiments.}
\label{tab:Lr-strategy}
\end{table}

\begin{figure}[!hb]
  \centering
  \includegraphics[scale=0.57]{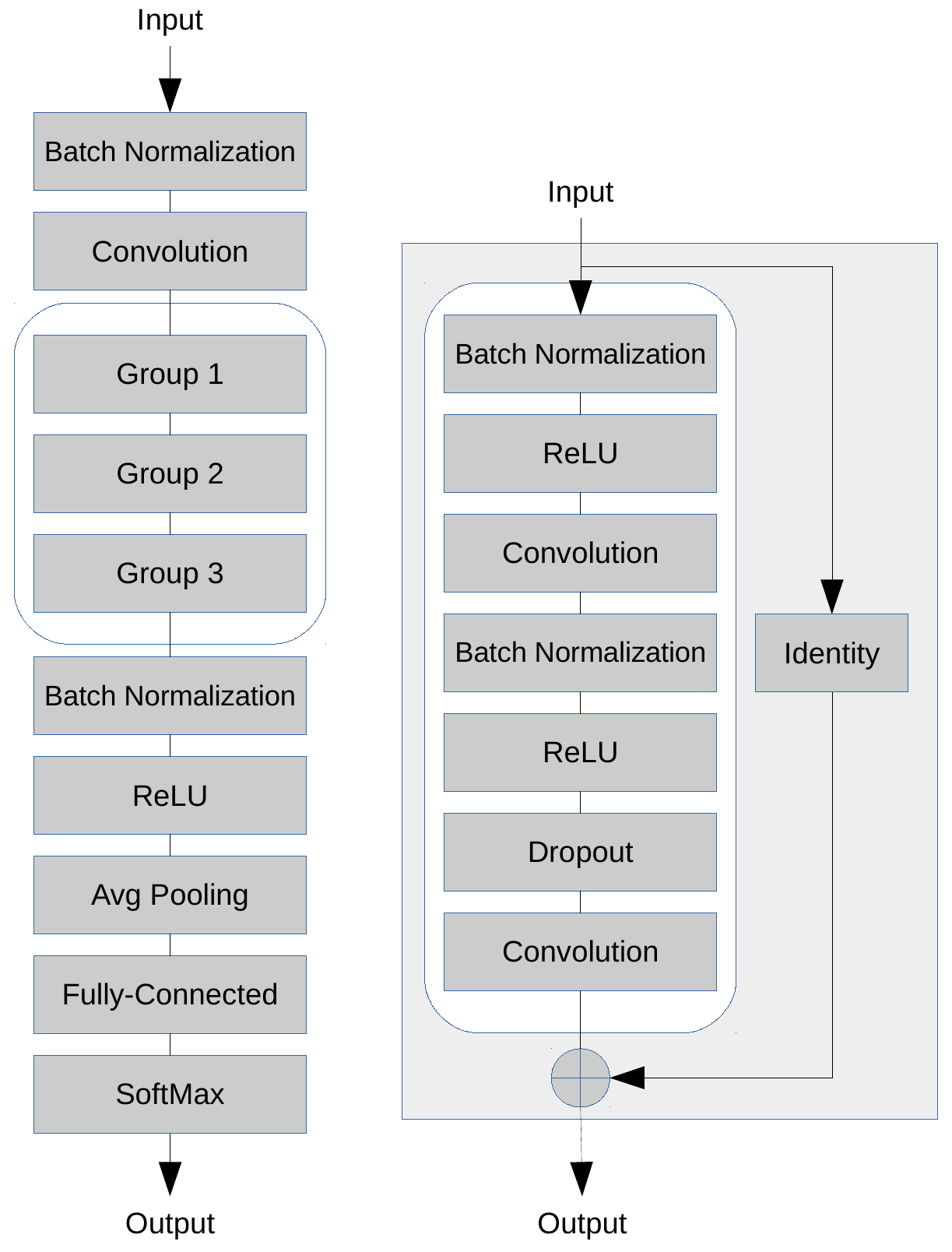}
  \caption{Structure of the proposed CNN model (left) and a typical convolutional block (right).}
\end{figure}

\begin{table*}[!t]
\footnotesize
\centering
\setlength{\tabcolsep}{1.0em}
\renewcommand\arraystretch{1.35}
\begin{tabular}{|l|cc|cc|}

\hline
Network & Depth & \#Parameters & CIFAR-10 & CIFAR-100 \\
\Xhline{3\arrayrulewidth}
NIN~\cite{Nin14}                        & -    & -     & 8.81  & -     \\ \hline
DSN~\cite{DSN15}                        & 3    & -     & 7.97  & 34.57 \\ \hline
All-CNN~\cite{Striving15}               & 9    & 1.3M  & 7.25  & -     \\ \hline
ResNet with Stochastic Depth~\cite{ResNetStoch16} & 110 & 1.7M & 5.23 & 24.58 \\
                                        & 1202 & 10.2M & 4.91  & -     \\ \hline
pre-act Resnet~\cite{He16}              & 1001 & 10.2M & 4.62  & 22.71 \\ \hline
Wide ResNet~\cite{WRN16}                & 40   & 55.8M & 3.80  & 18.30 \\ \hline
PyramidNet~\cite{PyramidNet16}          & 110  & 28.3M & 3.77  & 18.29 \\ \hline
Wide-DelugeNet~\cite{DelugeNet16}       & 146  & 20.2M & 3.76  & 19.02 \\ \hline
OrthoReg on Wide ResNet~\cite{OrthoReg16} & 28   & -     & 3.69  & 18.56 \\ \hline
Steerable CNNs~\cite{Steerable16}       & 14   & 9.1M  & 3.65  & 18.82 \\ \hline
ResNeXt~\cite{ResNeXt16}                & 29   & 68.1M & 3.58  & \textbf{17.31} \\ \hline
Wide ResNet with Singular Value Bounding ~\cite{SVB16} & 28  & 36.5M & 3.52  & 18.32 \\ \hline
Oriented Response Net~\cite{ORN17}      & 28   & 18.4M & \textbf{3.52}  & 19.22 \\ \hline
\Xhline{3\arrayrulewidth}
Baseline Wide ResNet                    & 28   & 36.6M & 3.62 & 18.29 \\ \hline
Volterra-based Wide ResNet              & 28   & 36.7M & \textbf{3.51} & \textbf{18.24} \\
\hline

\end{tabular}
\caption{Test set classification error results on CIFAR-10 and CIFAR-100, using moderate data augmentation (horizontal flipping, padding and $32\times32$ cropping).}
\label{tab:CIFARs}
\end{table*}

\subsection{Experimental setup}

In all our experiments we use Stochastic Gradient Descent (SGD) with momentum set to $0.9$ and cross-entropy loss, with a batch size of $128$, training our network for 220 epochs. Dropout is set to $0.3$ and weight initialization is done as in \cite{He15}. The learning rate and weight decay strategy used in the experiments is shown in Table \ref{tab:Lr-strategy}. For CIFAR-10 and CIFAR-100, the data-preprocessing operation applied to both train and test set's data, is subtracting the channel means and then dividing by the channel standard deviations, computed on the train set. We apply moderate data augmentation, using horizontal flipping with a probability of $50\%$ and reflection-padding by 4 pixels on each image side, taking a random crop of size $32\times32$.

\begin{figure}[b]
	    \hspace{-0.75cm}
	    \includegraphics[width=100mm]{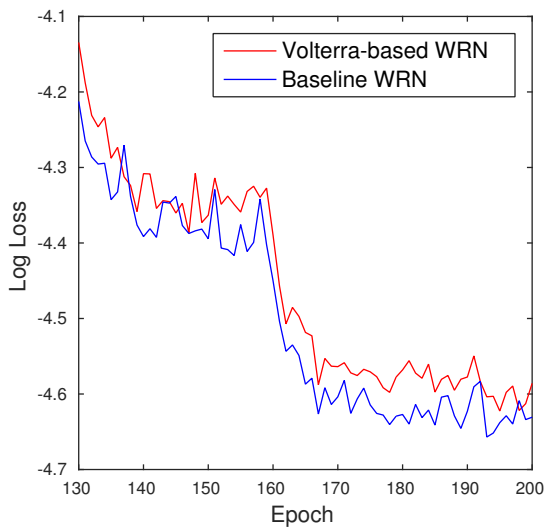}
	    \caption{Train loss on CIFAR-100.}\label{fig:CIFAR100-loss}
\end{figure}
\subsection{CIFAR-10 and CIFAR-100}

CIFAR-10 and CIFAR-100 datasets contain $60.000$ $32\times32$ RGB images of commonly seen object categories (e.g., animals, vehicles, etc.), where the train set has $50.000$ and the test set has $10.000$ images. CIFAR-10 has 10 classes and CIFAR-100 has 100 classes. All classes have equal number of train and test samples. In CIFAR-10, our Volterra-based Wide ResNet yields a test error of $3.51\%$, which shows an improvement over the $3.62\%$ error that we got using the baseline model, setting the state-of-the-art on this dataset. In CIFAR-100, our Volterra-based Wide ResNet yields a test error of $18.24\%$, which shows an improvement over the $18.29\%$ error that we got using the baseline model. Our results on CIFAR-100 are outperformed only by \cite{ResNeXt16}, due to the huge number of parameters their model makes use of. The features fed to the convolutional groups, when extracted by the non-linear convolution filters, make the network avoid overfitting. This can be inferred by the loss plot of our models on CIFAR-100, which is shown in Figure \ref{fig:CIFAR100-loss}. The Baseline Wide ResNet, although having constantly lower loss than the Volterra-based Wide ResNet, yields higher test error. Our Volterra-based Wide ResNets have only $0.05\%$ more parameters than the Baseline counterparts.
A summary of the best methods on these datasets is provided in Table \ref{tab:CIFARs}.


\begin{figure*}[!ht]
  \centering
  \includegraphics[width=1.05\linewidth]{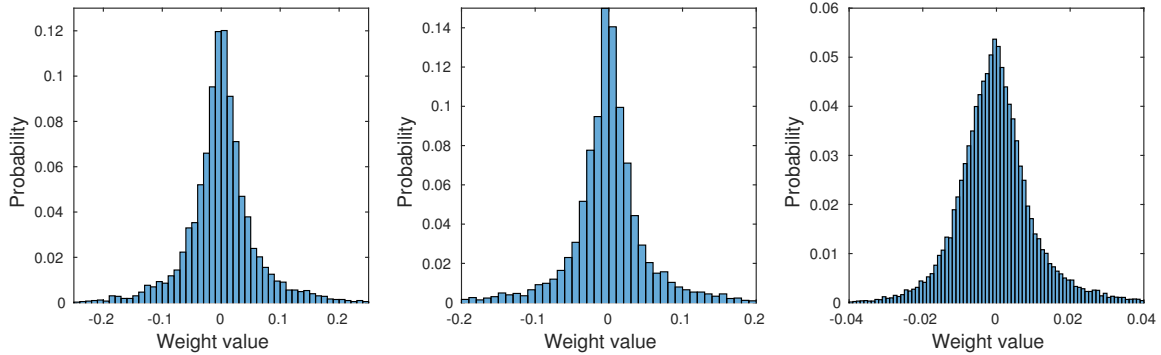}
\caption{Weight values of linear convolution filter weights (left), Volterra-based convolution first-order weights (middle) and Volterra-based convolution second-order weights (right).}
\label{fig:weight_hists}
\end{figure*}


\subsection{Weight visualization}
\begin{figure}[!hb]
\minipage{0.48\textwidth}
\hspace{-0.41cm}
\centering
  \includegraphics[width=0.25\linewidth]{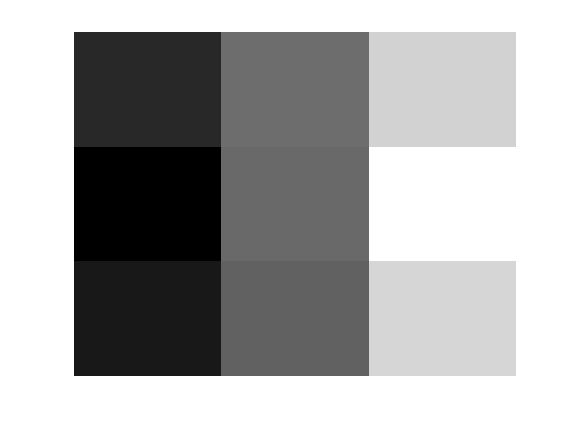}
\endminipage\hfill
\minipage{0.5\textwidth}
  \includegraphics[width=\linewidth]{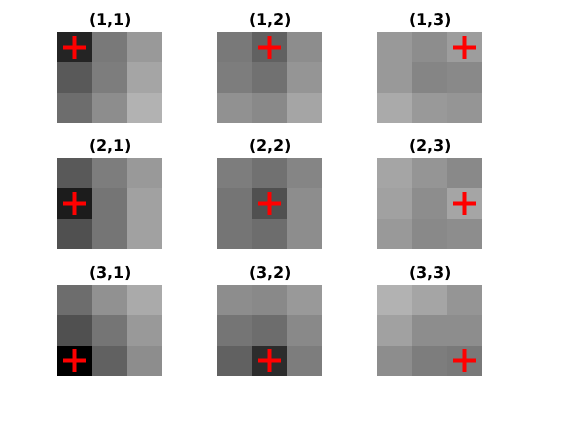}
\endminipage
\caption{Linear term and second-order multiplicative interaction weights of a Volterra-based $3\times3$ filter.}\label{fig:Volt_kern_vis}
\end{figure}	  
    To get an insight on what features do non-linear filters learn, we visualize their weights in a simple but efficient manner. For the linear term, the process is straightforward. For the second-order term, considering the weights $\vec{w}_2$ of each filter, we can create $n$ weight vectors $\vec{q}_i$, $\vec{q}_i=[w_2^{i1}, w_2^{i2}, ..., w_2^{in}]$. Reshaping each one of these vectors $\vec{q}_i$ into a $k_h \times k_w$ matrix, we can see the weights that correspond to the interactions between $x_i$ and all of the receptive field's elements. Figure \ref{fig:Volt_kern_vis} shows the weights of the linear term and the interactions captured by a second-order $3\times3$ filter, allowing us to explore their contribution to the response.
    Another issue, is the values that the weights of the non-linear terms settle to. We investigate these values, given the filters of the first convolutional layer of our Wide ResNet model, trained on CIFAR-100. The histograms of the weight values are shown in Figure \ref{fig:weight_hists}. The value distribution of the linear convolution filters' weights is similar to that of the quadratic filters' first-order weights. Also, the values of the quadratic filters' second-order weights have reasonably smaller standard deviation. 


\subsection{Response profiles}

\begin{figure*}[t]
  		\centering
  		\includegraphics[width=1\linewidth]{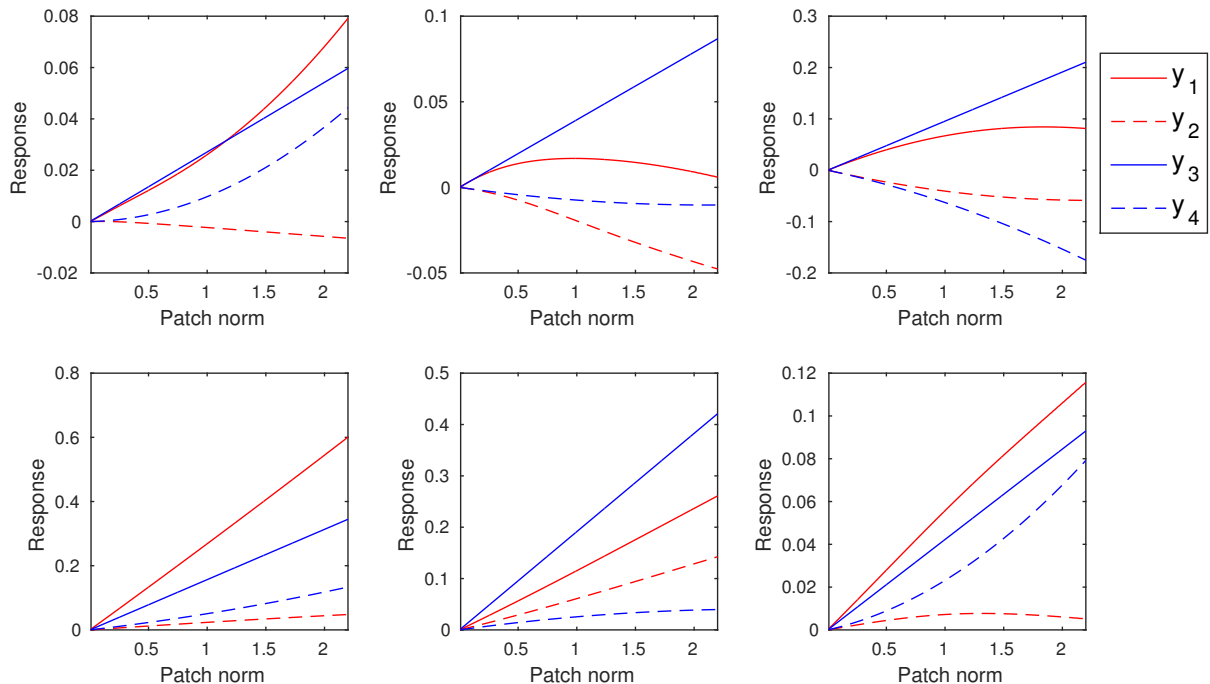}
  		\caption{Various cases of responses. Red line denotes the response $y_{1}$ of a Volterra-based convolution filter to its optimal stimulus $\vec{x}_o$. Dashed red line denotes the response $y_{2}$ of the linear subset of a Volterra-based filter's weights, to $\vec{x}_o$. Blue line denotes the response $y_{3}$ of the linear subset of a Volterra-based filter's weights, to $\vec{x}_l$. Dashed blue line denotes the response $y_{4}$ of a Volterra-based convolution filter, to $\vec{x}_l$.}
  		\label{fig:resp_plots}
\end{figure*}

   Following the methodology of \cite{Analysis06}, we use a set of Volterra-based filters of a Wide ResNet trained on CIFAR-100, to partly characterize their response profiles. Given the weights $\vec{w}_{1}$, $\vec{w}_{2}$ of a filter, we compute its optimal stimulus, $\vec{x}_{o}$, and the optimal stimulus of its linear term, $\vec{x}_{l}$, under the constraint that their norms are equal. Then, we compute four responses, as described in Table \ref{tab:resp_prof}, and plot them in Figure \ref{fig:resp_plots}. Comparing the various responses, we can infer that the properties of a linear filter with weights $\vec{w}_{1}$, can greatly change when it's extended to a second-order Volterra form by adding a weight set $\vec{w}_{2}$ with quadratic contributions. The response of a Volterra-based filter is quite different from the response of its first-order terms, proving that the second-order interactions contribute significantly to the functionality of a quadratic filter.
   
   Given the weight subset $\vec{w}_{1}$ of a Volterra-based filter, their optimal stimulus $\vec{x}_{l}$ has a standard pattern. As the norm of $\vec{x}_{l}$ takes values inside a bounded space, the way $\vec{x}_{l}$ varies is just a linear increase in all its intensity values, without altering its general pattern (i.e., all vectors $\vec{x}_{l}$ are parallel). However, this does not hold true for quadratic filters. As the norm of a Volterra-based second-order filter's optimal stimulus $\vec{x}_{o}$, takes values inside a bounded space, a rich variety of alterations can be observed in the elements of $\vec{x}_{o}$.

\begin{table}[!ht]
\centering
\setlength{\tabcolsep}{.3em}
\begin{tabular}{|c|c|c|}
\hline
\textbf{Stimulus} & \textbf{Filter} & \textbf{Response} \\
\hline
$\vec{x}_o$ & Quadratic ($\vec{w}_1$, $\vec{w}_2$)  & $y_{1}$ \\ \hline
$\vec{x}_o$ & Linear    ($\vec{w}_1$) 			   & $y_{2}$ \\ \hline
$\vec{x}_l$ & Linear    ($\vec{w}_1$) 			   & $y_{3}$ \\ \hline
$\vec{x}_l$ & Quadratic ($\vec{w}_1$, $\vec{w}_2$)  & $y_{4}$ \\ \hline
\end{tabular}
\caption{Stimuli, filter weight sets and filter responses.}
\label{tab:resp_prof}
\end{table} 


\section{Conclusion}
\label{Conclusion}

    The exploration of CNN architectures that are optimized for using non-linear convolution filters, is an open problem for biologically-inspired computer vision. Questions like \text{``}which is the ideal ratio between linear and non-linear filters in each convolutional layer?\text{''} and ``which properties prevail in the response profiles of each layer's non-linear filters?\text{''} are of great importance, to shed light in this hitherto unexplored category of filters. Any inference about the properties that are present to this group of quadratic filters, has the risk of being biased by the dataset used to obtain and observe them. This happens because the visual response profiles of the non-linear filters trained in the experiments, are constrained by the natural statistics of each dataset, as happens with the sensory system of primates, which adapts to its environment.

     Based on the research results of neuroscience that prove the existence of non-linearities in the response profiles of complex visual cells, we have proposed a non-linear convolution scheme that can be used in CNN architectures. Our experiments showed that a network which combines linear and non-linear filters in its convolutional layers, can outperform networks that use standard linear filters with the same architecture. Our reported error rates set the state-of-the-art on CIFAR-10, while being competitive to state-of-the-art results on CIFAR-100. We didn't apply our Volterra-based convolution to more layers, because our target was to demonstrate a proof of concept for the proposed method. Our claim was confirmed, as replacing only the first convolutional layer's linear filters with non-linear ones, we achieved lower error rates. Further testing quadratic convolution filters, is certainly an interesting direction for future work, to build better computer vision systems.

\section*{Acknowledgment}

   The research leading to these results has been supported by the EU funded project FORENSOR (GA 653355).



{\small
\bibliographystyle{ieee}
\bibliography{references}
}

\end{document}